# The role of media memorability in facilitating startups' access to venture capital funding

Toschi, L., Torrisi, S., & Fronzetti Colladon, A.

Toschi, L., Torrisi, S., & Fronzetti Colladon, A. (2025). The role of media memorability in facilitating startups' access to venture capital funding. Journal of Business Research, 200, 115627. https://doi.org/10.1016/j.jbusres.2025.115627

**The role of media memorability in facilitating startups' access to venture capital funding**

**Abstract**

Media reputation plays an important role in attracting venture capital investment. However, prior research has focused too narrowly on general media exposure, limiting our understanding of how media truly influences funding decisions. As informed decision-makers, venture capitalists respond to more nuanced aspects of media content. We introduce the concept of media memorability—the media's ability to imprint a startup's name in the memory of relevant investors. Using data from 197 UK startups in the micro and nanotechnology sector (funded between 1995 and 2004), we show that media memorability significantly influences investment outcomes. Our findings suggest that venture capitalists rely on detailed cues such as a startup's distinctiveness and connectivity within news semantic networks. This contributes to research on entrepreneurial finance and media legitimation. In practice, startups should go beyond frequent media mentions to strengthen brand memorability through more targeted, meaningful coverage highlighting their uniqueness and relevance within the broader industry conversation.

**Keywords**

Media reputation, Media memorability, Venture Capital, Semantic Network Analysis, Brand Knowledge, Semantic Brand Score.

# 1. Introduction

Technology-based startups suffer from the liability of newness and smallness (Freeman et al., 1983; Singh et al., 1986; Stinchcombe, 1965), and the chance of survival crucially depends on their ability to access resources from a diverse pool of external audiences (Gimenez-Fernandez et al., 2020). Due to the high development and scaling-up costs typical of high-tech companies, venture capitalists (VCs) represent one of the foremost providers of financial and managerial resources, networking, and other types of support (D. H. Hsu, 2004; Petkova et al., 2013). However, VCs have limited information on entrepreneurship projects' quality (Audretsch et al., 2009; Carpenter & Petersen, 2002; Leland & Pyle, 1977), and to make proper investment decisions, they analyze several signals of quality like some characteristics of the management team, the patents held by the investees, the attractiveness of the target market, the level of product differentiation, to cite some (Akerlof, 1970; Baum & Silverman, 2004; Franke et al., 2006, 2008; Grimpe et al., 2019; Haeussler et al., 2014; Hoenig & Henkel, 2015; Muzyka et al., 1996; Petty & Gruber, 2011).

In addition to this direct or indirect first-hand information, VCs rely on media as an informative source on startups' quality (Petkova, 2014). Media reputation, defined as the "overall evaluation of a firm presented in the media" (Deephouse, 2000, p. 1091), helps new organizations address the typical attention deficit problem, increases familiarity, and reduces the perception of risk, allowing startups to stand out from the crowd and become "identifiable in the minds of stakeholders" (Rindova et al., 2006, p. 66). Extant research has considered, in particular, *media visibility* – the number of news referring to a startup - as an important driver of media reputation (Kennedy, 2008; Pollock et al., 2008), which makes investors aware of a startup's existence (Berger et al., 2020; Petkova et al., 2013; Rindova et al., 2007). As earlier studies in social psychology suggest, repeated exposure to media may, in turn, affect audiences' attitudes (i.e., evaluation and predisposition) towards an object (Zajonc, 1968). In addition, scholars have also started to consider the role of media in framing identities in a positive or negative tone (Pollock & Rindova, 2003; Wei et al., 2017) by looking at *media favorability* (Bajo & Raimondo, 2017; Bednar et al., 2013; Bitektine, 2011; Deephouse, 2000).

These studies, however, have two key limitations. First, although they provide substantial evidence of media's influence, they often treat all audiences as a single, undifferentiated group. This overlooks the distinct characteristics that set stakeholder audiences, such as venture capitalists, apart from the general public, thereby failing to account for how different types of stakeholders interpret and respond to media content. Recognizing audience heterogeneity is relevant as audiences show different "taste positions" (G. Hsu, 2006, p. 423), operate with different institutional logics (Fisher et al., 2017), and give sense to signals they are exposed to in different ways (Berger et al., 2020, p. 202; Pontikes, 2012; Valkenburg et al., 2016). Second, despite the interest in the role of media having grown in management studies (Haack et al., 2021; Harmon et al., 2023; Oriaifo et al., 2020; Roccapriore & Pollock, 2023), they remain focused on media visibility and favorability, failing to dive into the characteristics of media. By characteristics, we refer to the content of the information delivered by the media. Our definition is taken from Pollock and Rindova (2003, p. 631): "we highlight the role of the media as an information intermediary and propose that characteristics of the information it provides serve as information stimuli that affect the formation of investors' impressions of firms". Not recognizing VCs as a peculiar stakeholder audience and failing to examine the role of characteristics of media limit our theoretical and practical understanding of the role of media in startups' VC funding because VCs are different from other types of audiences as they represent well-informed motivated experts, are specialized in evaluating startups (Baum & Silverman, 2004), sophisticated in interpreting information, for whom the influence of media reputation is not restricted to mere media visibility and favorability.

The analysis of the position of a startup's name in the semantic network of news, allows us to capture in more detail the relevance of the startup in the news discourse, offering additional cues for VC evaluations. The concept of a semantic network of news builds on Semantic Network Analysis. It refers to a set of words (nodes) found in news text, along with the connections between them, based on their co-occurrence (Bullinaria & Levy, 2007; Diesner, 2013). Compared to simply counting how often a startup name is mentioned, analyzing its position within the semantic network of news allows us to explore its associations with other terms in the discourse—its textual context. With this approach, we can capture how uniquely a startup is associated with certain concepts (i.e., the distinctiveness of its associations) and how well it connects otherwise unrelated or weakly linked concepts/terms (i.e., its connectivity) within the semantic network.

VCs are used to pay close attention to all the signals of a startup by adopting a complex, accurate, comprehensive, and detailed evaluation and decision-making approach based on information richness under uncertainty. It is, thus, interesting to dig deeper into the *characteristics* of media coverage to capture additional, finer-grained information about the position of a startup in the semantic network of media coverage that may affect the evaluation of VCs. Against this backdrop, we intend to investigate the following research question: *To what extent do news characteristics affect VC's propensity to invest in startups above and beyond media visibility and favorability?*

We capture the effect of the characteristics of the news through the concept of *media memorability*, which we define as the capacity of media to impress a startup name in the memory of VCs, thus increasing knowledge about it and its activities. The notion of media memorability contributes to this stream of research by digging into characteristics such as the distinctiveness of a firm name's associations and connectivity (measured by betweenness centrality of a firm name in the semantic network of news). In doing so, we draw on the concept of brand knowledge from the marketing communication literature (Keller, 1993), and we argue that media memorability results from the combination of three components. First, media prevalence is the frequency with which a startup's name appears in the news, which is supposed to create awareness about an organization and recall in the audience's memory (Keller, 1993). When a startup name is featured in the media, it can greatly increase its visibility and reach (Bruhn et al., 2012), with this exposure allowing more investors to become aware of a startup. Second, distinctiveness refers to the number and uniqueness of a startup name's associations, i.e., its links with concepts or information that are not very common in the news but are frequently associated with the startup name (Keller, 1993). Third, connectivity denotes the brokerage power of a company name resulting from the frequency with which it connects different concepts or topics in the media. Having a variety of less common associations that can also create a bridge between different concepts is advantageous in establishing distinctive links between the brand and other memory nodes in investors' minds. This, in turn, can increase the likelihood of the startup name being remembered, as it constitutes a node in the memory of investors that can be frequently activated by exposure to other linked memory nodes (Bryant & Oliver, 2009; Krishnan, 1996).

By combining three dimensions of media – prevalence, distinctiveness, and connectivity we extend extant research on media reputation by highlighting dimensions of media coverage that have not been examined before. In doing so, we address a relevant gap in the literature since, as mentioned before, venture capitalists are sophisticated decision-makers who can systematically screen the media and analyze the signals of unobservable startups' quality provided by the distinctiveness and connectivity of startups' brand names in the semantic network of news.

Empirically, we study the effect of media memorability on the VC decision process in the micro and nanotechnology (MNT) sector, exploiting information on VC investment decisions and media exposure on 197 MNT startups founded from 1995 to 2004 in the United Kingdom (UK). The industry, the geographical area, and the period chosen are an ideal setting for conducting our study. The MNT was an emerging sector, characterized by complex technologies and high uncertainty, two conditions that magnify the value of credible external signals such as those provided by media. In the period under investigation, the MNT sector was, indeed, in its emerging phase of development, characterized by intense ferment in terms of new firm formation, rapid technological change, and an unclear regulatory setting (Aitken et al., 2006; Linton & Walsh, 2004; Rambaran & Schirhagl, 2022). VCs are 'naturally' attracted by businesses and tech fields with high opportunities and uncertainty, like the MNT. However, at the same time, the high uncertainty and information asymmetries surrounding the MNTs make media an extremely relevant informative intermediary for VCs' investment decisions. The UK, moreover, is one of the most vibrant VC markets in Europe and one of the most attractive places for entrepreneurship and innovation, with a large number of startups utilizing and developing MNTs (KTN, 2010; Mason & Harrison, 2002; Talebian et al., 2021). Finally, the UK is one of the most established and influential media industries globally. The national diffusion and high visibility of many media outlets heighten the potential impact of memorable news exposure on investor attention. During the pre-digital period, traditional media played a particularly influential role in shaping reputational signals, as startups had limited access to alternative channels to manage their visibility.

By analyzing the characteristics of the news, we find that higher levels of media memorability correspond to a higher probability of being financed by VCs. We, thus, contribute to the discourse on the association between startups' media reputation and VC funding by introducing a novel perspective based on the analysis of dimensions conveyed by media that remained unexplored in earlier works. More specifically, our work contributes to VC decision-making research by highlighting dimensions of media that VCs can use as proxies for otherwise unobservable startup characteristics. Moreover, our findings point to media memorability as an important dimension that startups could take into account in their sensegiving activities, aiming to attract the attention of external stakeholders.

This study is organized as follows. First, we present our theoretical background and hypotheses. Then, we illustrate the methods (data, variables, and empirical model) before presenting our results and robustness checks. In the discussion, we elaborate on our main contributions, acknowledge limitations, and provide suggestions for future research.

## 2. Theoretical background and hypotheses development

### 2.1. The role of media as key information intermediaries

Media are a crucial intermediary able to attract public attention to firms' activities (Reese & Shoemaker, 2016), enable the acquisition of knowledge useful to different stakeholders (Dorobantu et al., 2024), reduce information asymmetries (Fang & Peress, 2009; Li et al., 2023), affect stakeholders' evaluations of firms (Pfarrer et al., 2010; Pollock et al., 2019), increase transparency (Hammami & Hendijani Zadeh, 2019), or influence competitive dynamics (Tan, 2016).

Recent studies have documented the importance of media coverage (Bettinazzi et al., 2024; Dyer et al., 2022) in public perception and financing of new ventures. For instance, Guldiken et al. (2017) apply signaling theory to explore the influence of media coverage on IPO stock performance. Their findings reveal that credible financial media coverage of an IPO firm has a significant effect on its stock price. In contrast, uncertainty in the tone of news articles regarding an IPO firm negatively impacts its stock price. Naumovska and Harmon (2024) investigate how media coverage in leading US business media shapes the financing outcome of founders in special purpose acquisition companies (SPACs). This work clarifies that investors have limited access to information about a startup's internal operations, intellectual property, human capital, and relationships with suppliers, customers, or partners. Thus, SPACs can reduce information asymmetry by communicating with investors through formal channels like IPO prospectuses filed with the Securities and Exchange Commission. These communications are legally binding, which creates incentives for truthful communication. The authors do not find evidence of a significant effect of media coverage on IPO outcome in addition to the information reported in the IPO prospectuses. Mochkabadi et al. (2024), in the UK context of equity crowdfunding, analyze whether the relationship between the distinctiveness of new ventures (i.e., level of innovativeness) and their performance (i.e., resource acquisition) is influenced by external endorsements such as prominent media coverage in top television broadcasters views and top newspapers. They find a positive and significant moderation effect of media coverage on the relationship between incrementally innovative ventures and equity crowdfunding (number of investors and amount raised). Instead, media coverage has an insignificant moderating effect on the crowdfunding outcome of radically innovative and non-innovative ventures. Media also play a key role in shaping public perceptions of entrepreneurship. Schmid and Welter (2024), focusing on media coverage in top German newspaper outlets from 2000 to 2021, examine how news uses labels, language, and linguistic style, in addition to salient socio-cultural content aspects legitimized by media sources. The paper draws on agenda setting, (media) legitimacy, and narrative research that provide valuable insights into how volume, frames, and language in media shape public perceptions. However, the study does not investigate the effect of these dimensions of media coverage on stakeholders' (investors') decision-making. Similarly, Nadin et al. (2020)

are interested in the role media play in producing cultural and social meanings, particularly in representing masculinity and femininity in entrepreneurship.

In addition to traditional media, the entrepreneurship literature has highlighted the importance of modern and alternative channels like digital and social media (Olanrewaju et al., 2020; Secundo et al., 2021) to create and connect startups directly with their community (Fisher, 2019), building emotional links through multimodal communication which combines images and words (Garud et al., 2014; Puschmann & Powell, 2018; Roccapriore & Pollock, 2023), develop identity and legitimacy (Etter et al., 2018; Glozer et al., 2019; Horst et al., 2020), and influence attitude and behaviors of investors (Bayar & Kesici, 2024; Jin et al., 2017; Nguyen et al., 2020; Wang et al., 2024).

In the literature on startup financing, researchers have increasingly addressed the role of digital and social media as a channel that may help reduce the information asymmetry between new ventures and investors. For example, Aggarwal et al. (2012) found that online reviews and judgments might affect the likelihood that a firm receives VC financing. Similarly, Jin et al. (2017) have shown that social media activities and strong Twitter influence (followers, mentions, impressions, and sentiment) increase the likelihood of startups receiving VC financing. Interestingly, Wang et al. (2024) found that the effect of social media on VC financing is heterogeneous. In particular, using Twitter increases the likelihood of securing VC financing for firms founded by women and individuals with limited social capital (i.e., individuals with weak connections to the investor network). Moreover, Lin et al. (2013) showed that the online friendship of borrowers acts as a signal of credit quality and increases the chance of successful funding in the online market for peer-to-peer (P2P) lending.

This overview suggests that despite the popularity of new digital and social media, traditional media are still an influential, reliable information intermediary in the management field, in general, and in the evaluation of technological ventures and the financing of innovation. Like traditional media, social media provide investors with useful information about investment opportunities and help them reduce the information asymmetry about startups' quality. However, unlike traditional media, social media are less likely to offer the same degree of legitimation as professional media since social media users do not have to ensure the level of information accuracy of traditional media (Veil et al., 2012).

## 2.2 The role of media memorability in the evaluation of startups

The characteristics of media coverage may provide useful information to evaluate startups better, in addition to the volume and tone of news. The structure of news provides information about the startup name's associations with other concepts in the media discourse, thus revealing the position and strength of the startup in the semantic network of news. A stronger, more persistent impact increases the perceived value of a new organization in the eyes of expert decision-makers like VCs, who need fine-grained information to apply decision practices based on rich connections among concepts (Shepherd & Zacharakis, 2002). In other words, a deeper analysis of media news concerning a firm may tell about *media memorability*, which we define as the capacity of media to impress a startup name in the memory of stakeholders. Building on marketing communication studies and semantic network analysis, we argue that media memorability may uncover features of media reputation not captured by the notions of media visibility and media favorability.[1] Our main theoretical point is that memorability is a dimension resulting from the combination of the frequency of a brand citation in the news, the distinctiveness of brand associations, and the connectivity of a brand in the discourse where it is positioned[2].

---

[1] Unlike social networks connecting people to people, semantic networks connect words to words, based on their co-occurrence (Hansen et al., 2020).

According to the associative network memory model (Collins & Loftus, 1975; Keller, 1993; Krishnan, 1996), memory is a repository of thoughts, mental constructs, or cognitions. Information stored in memory can be represented as nodes connected by associative pathways, with links reflecting how frequently two nodes are co-activated when processing incoming information. In a network model, the priming mechanism occurs when a node (i.e., a piece of information) is temporarily accessible to the mind by stimuli from the external environment or the individual's cognitive system, and activation spreads to related nodes (Ewoldsen, 2020). Priming is, thus, a procedure that increases the accessibility of a concept in memory. In the media setting, priming occurs when exposure to news affects individuals' later behavior or judgment (Ewoldsen & Rhodes, 2019). Nodes that are activated frequently develop chronic accessibility, which refers to concepts that are always highly accessible from memory. Chronically accessible concepts are characterized by a low activation threshold. Many brand decisions are triggered by a similar mechanism, which is the accessibility of information in memory (Walvis, 2008), and are driven by long-term memory.

Researchers in cognitive psychology confirm that the activation level of a node is durable (Loftus & Loftus, 1980), although the strength of its associations dissipates over time if no additional activation occurs (Bryant & Oliver, 2009). Thus, if no 'reinforcement' occurs, even chronically accessible concepts become less accessible over time (Grant & Logan, 1993). The intensity of priming, which depends on the frequency of exposure to a stimulus (e.g., news about a brand), produces stronger, more lasting priming effects (Bryant & Oliver, 2009; Higgins et al., 1985). In our context, this suggests that media visibility, i.e., the frequency of a startup name's associations in the news, may induce a persistent receptiveness of the brand in stakeholders' memory, thus increasing awareness and media memorability. The cognitive psychology perspective on priming has been adopted by marketing studies that view brands as networks of associations with a name in people's brains. In this view, brands are "pieces of information, meanings, experiences, emotions, images, intentions, etc. interconnected by neural links of varying strength" (Walvis, 2008, p. 180).

---

[2] Our analysis focuses on startup names, which we consider as 'brands' or distinguishing names intended to identify the goods or services of a firm (Aaker, 1991). We did not analyze individual product brands, as this information was often unavailable, and product names/brands are frequently subject to change for startups. In the news we analyzed, it is common to see statements like "Startup Alpha develops a new drug to cure skin cancer", where the product name is not highlighted. In other high-tech industries like software, firms are particularly concerned with their firm-level brand and use trademarks to communicate corporate brands and acquire market visibility rather than to promote specific products or services (Giarratana & Torrisi, 2010, p. 97).

Another component of media that likely increases the effect of memorability on stakeholders' decision-making is the distinctiveness of associations. Brand equity studies suggest that, besides awareness, the knowledge about a brand depends on the brand image reflected by the brand associations (Keller, 1993). Brand associations represent the information connected to various dimensions of a brand in consumers' memory. They shape the brand's meaning by encompassing attributes that consumers believe the product or service possesses, the benefits or value they assign to those attributes, and their overall evaluation of the brand. More numerous, stronger, and more distinctive brand associations (i.e., associations that a brand does not share with other brands) represent an important dimension of brand image and are deemed to provide a firm with a sustainable competitive advantage or unique selling proposition (Hoeffler & Keller, 2003; Keller, 1993). Unfortunately, the empirical evidence about the benefits of distinctive associationss is inconclusive. Lab experiments conducted by Keller (1987) found that the number of brand ads in the same product category (a proxy for shared associations) negatively affects consumer memory and brand evaluation. Krishnan (1996) also found a positive association between brand equity and unique brand image (an extreme case of distinctiveness), while Romaniuk and Gaillard (2007) did not find any significant association between brand uniqueness and brand performance (brand preference and brand market penetration). Moreover, Grohs et al. (2016) found that the uniqueness of an individual's brand associations has no clear effects on brand strength. Based on a sample of consumer goods, they found both cases of brands with several unique associations and high brand strength and cases with a small number of unique associations and high brand strength.

Finally, an additional dimension critical to creating memorability is connectivity, which is how embedded a startup's name is in the broader discourse. Being central and not peripheral to the overall discussion means being able to connect different discourse topics presented in the news, which contributes to the generation of memorability. To understand the implications of connectivity for media memorability, we build on the notion of brokerage (a broker who stands between two disconnected parties), taken from social network analysis (Burt, 2004). In this stream of research, individuals whose social connections bridge the gap between groups benefit from access to a variety of information and their experience in translating information between groups. These individuals own social capital that gives them a competitive advantage in identifying and developing promising opportunities (Burt, 2004). Brokerage could create value on various grounds. First, individuals familiar with activities in two separate groups can transfer best practices from one group to another after translating these practices into a language the receiving group can understand. Often, group members do not recognize the value that other groups' practices may have for their group's operations. Second, brokers develop the ability to find new ideas by combining elements from two distinct groups. In line with these features of brokers, Burt (2004) found evidence that managers whose networks span structural holes are more likely to contribute new ideas, discuss them with colleagues, and receive a positive evaluation of their ideas. Works focusing on different settings, such as digital business ecosystems (Isherwood & Coetzee, 2011) and the World Wide Web (Gloor et al., 2009), have analyzed brokerage by using the notion of betweenness centrality from social network analysis (Varlamis et al., 2010). Betweenness centrality measures the extent to which a node lies on the shortest paths that connect other pairs of nodes in a network and can be calculated as the fraction of the shortest paths between node pairs that pass through a third node (Brandes, 2001). A high level of betweenness implies that, by facilitating interactions, a central node can control information flows between unconnected nodes (Isherwood & Coetzee, 2011). The notion of betweenness centrality is relevant in our setting, where words (nodes) are connected to other words, since a central position in the semantic network of news can allow a brand to stand out from the mass of brands covered in the media discourse, thus obtaining a stronger effect on the long-term memory of stakeholders.

To summarize, we suggest that the three components of media memorability work by activating the chronic accessibility of a startup name in the memory of stakeholders. A startup whose name is frequently mentioned in the media, is embedded in a rich and distinctive discourse, and connects different topics, has a greater competitive advantage over other startups in terms of attracting stakeholders' attention and revealing the prominence of the startup in the semantic network of news.

## 2.2. VCs as well-informed decision-makers

Stakeholders do not share a uniform perspective in evaluating firms; their views are shaped by unique lenses influenced by their perceptions and intentions (Zhao et al., 2017). In this regard, Pontikes (2012) offers a broad distinction, categorizing stakeholders into market-takers (consumers) and market-makers (VCs). This classification is particularly salient in the finance of technology startups. In evaluating innovative companies that develop offerings that are able to redefine the market structure, consumers tend to rely on conventional categories. They are thus unlikely to tolerate offerings that diverge from the known and are difficult to classify. Conversely, VCs prioritize novelty, showing a preference for ambiguous classifications and innovative offerings.

Given the VCs' positive attitude toward organizations with an ambiguous identity, it is interesting to see how a startup identity conveyed by the media influences VC investment decisions. This effect depends on the characteristics of VCs, such as their mental construal, or mindset underlying their investment decision process (Trope & Liberman, 2010; White et al., 2011), which can be described as follows.

First, VCs are expert investors in startup evaluation (Petty et al., 2023). Expertise is typically defined as having high levels of knowledge in a specific domain (Johnson, 2013). Cognitive research suggests that expertise influences how information is processed (Maheswaran et al., 1996). In particular, experts "differ in the amount and types of information they selectively consider" compared to non-experts (Alba & Hutchinson, 1987, p. 419; Gibson, 1994). Expertise increases the ability to extract information from external sources so that, when making evaluations, experts can recognize and rely on a broad and detailed pool of informational cues that non-experts cannot appreciate (Boulongne et al., 2019). In other words, experts "make use of attributes that are ignored by the average person" (Rosch et al., 1976, p. 430). Thanks to their expertise, VCs have a superior capacity to find and interpret external information that can be useful in estimating the potential value of the investment in a startup.

Second, VCs are motivated audiences, as they are interested in obtaining tangible (financial) gains from their investments. Cognitive psychology research suggests that individuals tend to dig deeply into areas that spark their interest. "Interest is a source of intrinsic motivation that fuels learning" (Johnson, 2013, p. 335), and this nurturing mechanism leads to the construction of considerably richer mental approaches. This means that motivated individuals not only scrape the surface but spend a great deal of time observing, reasoning, and analyzing, and the amount of information they can store in memory allows them to identify objects that less motivated and interested individuals fail to identify.

Third, VC activity is inspired by evidence-based management principles as they typically incorporate the best available scientific evidence to support their decision-making (Briner et al., 2009). Specifically, adopting a "scientific approach" (Felin & Zenger, 2015) in contexts characterized by high uncertainty, like investment in technology startups, has been documented to improve the ability to reduce "the odds of pursuing projects with false positive returns, and raises the odds of pursuing projects with false negative returns" (Camuffo et al., 2020, p. 564). A scientific approach is characterized by the formulation of conclusions based on objective and rich data. It results from a systematic process that is done purposively, selectively, and in-depth and relies on reasonable facts (Hassan & Hanapi, 2013). As such, decision-makers adopting this approach do not dwell on first impressions but are rather guided by rigor and concrete evidence to better understand the object under investigation (Camuffo et al., 2020). Regarding VCs, this implies adopting a "conscientious, explicit, and judicious" process to evaluate potential investments (Briner et al., 2009, p. 19). VCs are used to operating in an information-rich and uncertain environment that strains their information-processing capabilities (Shepherd & Zacharakis, 2002) and spur directing their attention primarily toward key dimensions, creating categories of information, and developing strong links between concepts and adopting decision-making practices that utilize these rich connections (Shepherd & Zacharakis, 2002). An increase in the amount and quality of relevant information allows decision-makers to make more accurate judgments (Zacharakis & Shepherd, 2001) and better decisions (Camuffo et al., 2020).

Given their expertise, sophisticated mindset, and evidence-based decision-making, it is, thus, plausible that VCs will select finer-grained information and rich cues that less sophisticated audiences could overlook. Media memorability, operating under the same logic of offering information richness, is thus supposed to play a particularly relevant role in the VC decision-making process. With the notion of media memorability, we put media visibility, a measure of awareness and prominence already explored by earlier studies, in a broader setting where media visibility contributes along with other dimensions (distinctiveness and connectivity) to affect the accessibility of information in the memory of VCs (Keller, 1993). Therefore, we expect that:

*H1: Media memorability is positively associated with the likelihood of receiving venture capital funding.*

## 2.3. The moderating role of media favorability

Our definition of media memorability does not consider the tone of a firm's media coverage. The reason for keeping the tone of a firm's media coverage separated from media memorability is that the latter does not rely on explicit subjective evaluations. The three components of media memorability reflect the frequency and positioning of a firm's brand name in the media discourse about the firm, which reflects nonevaluative recognition of a firm's brand name (Wei et al, 2017). Instead, favorability reflects evaluative judgment as it indicates the tone, i.e., positivity or negativity, of the discourse about a brand (Deephouse, 2000). Indeed, one can remember the name of a company very well because of the frequency, distinctiveness, and connectivity of its name in the semantic network of media coverage, regardless of whether one has a good or bad opinion about it. Integrating subjective evaluations captured by the tone of the news into the notion of media memorability would introduce a conceptually confounding factor into a dimension that represents different important objective factors resulting from the news analysis. Therefore, the separate effects of both constructs should be considered while evaluating the impact of different dimensions of media coverage on VC decisions. That said, the tone of media coverage may play an important role in forming media reputation as it can affect the overall salience of the objects in the news (Kiousis et al., 1999). Therefore, in our analysis, we consider this dimension as a separate explanatory factor.

Why is the tone of media coverage important in our setting? Deephouse (2000) has found that media favorability has a positive impact on the profitability of established organizations, while other studies (Aggarwal & Singh, 2013; Bajo & Raimondo, 2017; Fombrun & Shanley, 1990) point to the role of the tone of media news in a new firm's capacity to attract resources. The tone of media coverage - positive, neutral, or negative - reflects (or may affect) various stakeholders' beliefs about a firm's future performance. Rindova and Fombrun (1999) have discussed the role of reputational ratings in reducing uncertainty about firms' likely behaviors or future performance levels. They also mention the role of institutional intermediaries, like the media, in transmitting and magnifying information about firms and as an intermediary that helps the constituents categorize firms and evaluate their ability to deliver value (Rindova & Fombrun, 1999, p. 700). Moreover, positive media favorability is difficult for competitors to replicate, while the complexity of factors leading to a favorable reputation protects the firm's competitive advantage and future profitability from the threat of imitation. In addition, a favorable reputation is not tradable (Deephouse, 2000), which imbues it with strategic value (Dierickx & Cool, 1989). Furthermore, a positive reputation gives rise to a classical gain spiral (Boyd et al., 2010). In markets where it is difficult to distinguish good projects from "lemons", a favorable media reputation allows firms to enhance their access to capital markets (Beatty & Ritter, 1986) as it is a signal that VCs can use to reduce uncertainty (Akerlof, 1970; Spence, 1973).

Media favorability adds an affective dimension to the information about the characteristics that describe an object (McCombs & Reynolds, 2008; Zhang, 2018). These conveyed feelings cause information to "become more salient, to stand out more, arouse more interest, cause deeper processing and greater learning of congruent material" (Bower & Cohen, 1982, p. 291). Thus, we suggest that favorability reinforces the effect of media memorability on the evaluation and behavior of VCs because it magnifies a firm's distinctive, central position in the memory of its stakeholders. Moreover, increased memory for positive stimuli leads to more positive brand evaluations, such as a higher attitude toward the brand (Keller, 1987; Lee et al., 2016). From all the insights, in our context, we expect that when a startup has become memorable, and the general impression of the firm is favorable, VCs will be more prone to provide VC financing. Thus, we suggest:

*H2: The effect of media memorability on the likelihood of receiving venture capital funding is higher when media memorability is combined with high media favorability.*

## 3. Data and methods

### 3.1 Research setting: startups in the micro and nanotechnology sector in the UK

To analyze the effect of media reputation on access to VC funding, we chose startups founded from 1995 to 2004 that operated in the micro and nanotechnology (MNT) sector in the United Kingdom (UK). MNT includes a wide range of advanced techniques used to study systems, with dimensions ranging from several micrometers to a few nanometers. MNT is a breakthrough discovery, which, like information and communication technology or biotechnology, can change our everyday lives and offer significant technological opportunities and economic value (Rothaermel & Thursby, 2007). It is thus a valuable context for VC investors who search for valuable projects to finance and obtain potentially highreturn from their investments. As an interdisciplinary science, it can have numerous applications ranging from food, cosmetics, and agriculture to medical products and semiconductors (UK Nanotechnologies Strategy; Small Technologies, Great Opportunities, 2010). The world market for MNT-based products was valued at USD 11.99 billion in 2022, with an expected increase at a compound annual growth rate of 18.05% from 2023 to 2032 (*Nanomaterials Market Size, Share, and Trends*

*2024 to 2034*, 2024). The field's importance is further signaled by the growth of private and public investment directed toward MNT to support the sector's scientific and technological development. We focused on the period 1995-2004, which represents an important stage of the development of nanotechnology[3]. After two pivotal enabling technological breakthroughs that propelled nanotechnology to the forefront of physical science research (Rothaermel & Thursby, 2007) – the Scanning Tunneling Microscope (STM) in 1981 and the Atomic Force Microscope (AFM) in 1986 – nanotechnology was still an emerging field until the early 2000s. The number of scientific publications and patent applications in the field remained limited until the mid-1990s and increased dramatically afterward (Bozeman et al., 2007; Rothaermel & Thursby, 2007). In this period, the nanotechnology sector exited the incubation phase and began to capture public attention, subsequently evolving from basic research to applied research and commercial applications (Aitken et al., 2006; Linton & Walsh, 2004; Munari & Toschi, 2011). The focus on the nanotech field's early phases of development is interesting and relevant for at least two reasons. First, by analyzing a startup's media reputation and VC funding in an emerging, fast-growing, and yet fluid field characterized by rapid technical change and high uncertainty about the market potential (Granqvist et al., 2013; Grodal, 2018), we highlight the role of an important dimension that studies on industrial evolution and the diffusion of innovation have not deeply investigated (Abernathy & Utterback, 1978). Second, while the technological and industrial landscapes have changed since the second half of the 2000s, the nanotech field is still far from a maturity stage, as demonstrated by the high level of technological, regulatory, and market uncertainty. The rate of new firm formation is still high, and no rapidly growing firm has emerged yet, unlike other fields like digital technology and electric cars. Technological uncertainty is due to factors such as the difficulty in scaling up a process from the lab to the industrial operation, a lack of understanding of the interaction of nanomaterials in vitro and in vivo, inadequate knowledge of bioaccumulation in target organs, tissues, and cells, and information about their biocompatibility. The lack of technology transfer protocols or requisites for regulatory approvals hampers the transfer from the R&D labs to commercial products, thus contributing to market uncertainty (Rambaran & Schirhagl, 2022) .

---

[3] The high fluidity of the nanotech field probably explains why recent studies that investigate the nanotech field's social and symbolic boundaries over time (Grodal, 2018) have also used data in the period from the early 1980s to 2005, on the assumption that the observed regularities are still valid.

Although the use of MNT in today's marketplace continues to expand and attract private investment, financial resources are still inadequate because of "insufficient funding opportunities to engage in research that has the potential for commercialization" (Rambaran & Schirhagl, 2022, p. 3673). Specific barriers have prevented VC funding from matching the pace of scientific advancements. First, nanotechnology investments often lead to incremental improvements and generally offer modest economic and societal benefits. This scenario makes it difficult for MNT-based companies to grow large and make high profits and, consequently, attract VC funding. Third, because MNT is often a small piece of the final product, it cannot usually fully alter a current technology or create a new product, further discouraging VC investments. Finally, besides technological and regulatory uncertainty, uncertainties in intellectual property rights create another barrier to VC involvement in MNT. As previously stated, MNT is an interdisciplinary technology; thus, identifying appropriate patent classes can be difficult. This lack of standard definitions has hampered patent offices' evaluation procedures, leading to overlapping and conflicting claims among applicants. As VCs generally consider the ownership of patents as a signal of quality (Mohammadi & Khashabi, 2020), an unclear scenario in terms of technological protection makes them hesitant to take on the risk that an investment may not be fully exploited in the market. In this scenario, characterized by high uncertainty and risk for investors, understanding how media can help VCs run the race is a valuable exercise. In the observed period of nanotech evolution, traditional media were an important source of reputation for small companies and a relevant additional informative source that investors could use to address uncertainty and uncover valuable investment opportunities. It is reasonable to assume that the persistently high uncertainty makes media reputation a relevant source of information for professional investors.

Lastly, our focus on the UK industry offers all the factors needed to investigate our research question. First, the UK has a strong technical base and a good track record of domestic startups in the field. Recent data show that the UK and Germany are the top European contributors to product specifications in the MNT industry (Inshakova et al., 2020; Talebian et al., 2021). Moreover, UK MNT firms operate in all stages of the supply chain (KTN, 2010) and rank third among the world's top suppliers, after their competitors based in the United States and China (UK Nanotechnologies Strategy; Small Technologies, Great Opportunities, 2010). Second, the UK represents Europe's most developed VC market, with increased investments in the years under investigation. Indeed, although the 1980s were the flourishing years for the VC industry in the US, only during the 1990s was there enormous growth in VC deals in the UK, while, in other European countries, the market was still in its infancy (Mason & Harrison, 2002). The amounts raised by UK investors increased from less than £0.5 billion in the early 1990s to £9 billion in 2000 (BVCA Report on Investment Activity 2000, 2001). Third, the UK has a well-established media industry with several influential national, regional, and local newspapers and a strong tradition of journalism. Moreover, in the period investigated, this market was still dominated by print circulation with about 16 million copies sold daily, placing it among the higher ranges in Europe, although digital media was emerging as a new paradigm. These reasons allow us to investigate a context with a position of primacy in the development of MNTs, where an effective market of equity-based investment can support such innovations.

## 3.2 Data collection

We collected data from various sources. Firstly, we started with the Industrial Map of UK MNT, which identified 372 firms operating in the MNT sector in the UK by 2004. Among these companies, 201 were classified as startups. The Industrial Map of UK MNT was collected by the MNT Network and the UK Department of Trade and Industry in 2004, to provide an initial mapping of the activities performed by UK firms in the MNT sector. The Map contains basic firm information, including founding year, region, and corporate origin. Secondly, we used Companies House to check for the startups' name changes and status. Thirdly, we gathered VC funding data from Thomson One. Through Thomson One, we also identified 9 additional VC-funded startups operating in the MNT sector, three of which were founded in 2005-2007. We, thus, included only 6 of these additional observations in our final sample. Moreover, we used startup director data to operationalize variables at the management team level. In particular, the director list of each startup was drawn from the FAME dataset. FAME is a database containing comprehensive economic and legal information on companies in the UK and Ireland (Forti et al., 2020). For each company, FAME provides the current and previous list of directors, alongside each director's title, appointment date, resignation date, and birthday. We then used Company Director Check to collect information on each director's previous occupational experience background, while data on directors' publications were collected from Scopus. We excluded 10 startups from our sample for which FAME did not report any information. Lastly, we collected European patent applications[4] from QPAT, Questel-Orbit's Internet database providing the full text of patents from

---

[4] We used patent applications, rather than granted patents, in line with the extant literature which suggests that the two capture different aspects of patenting activity. Granted patents measure the ability of a startup to protect their innovations by excluding other firms from the use of an invention and, thus, they support the appropriation of returns from innovation and allow the creation of a competitive advantage (Hall et al., 2005, 2007; Hall & Ziedonis, 2001). Conversely, patent applications measure the extent to which companies put an effort into technological development and represent a signal not only of future performance, but

1974 to the present for 95 patent offices (Useche, 2014), including European patent applications published by the European Patent Office. After these steps, the final sample includes 197 startups founded in 1995-2004, 66 of which received at least one round of VC funding by December 31, 2010 (the last observation year), while 131 are non-VC-funded companies. As our sample comprises startups founded between 1995 and 2004, our observation period ends in 2010, which allows us to have a proper time window (at least five years) to trace VC funding for companies founded in 2004. For this sample of companies, we finally collected data on media exposure by gathering news from LexisNexis Academic. This database collects articles from different types of outlets, such as media, press releases, and academic journals. We filtered on the categories "Newspapers (National, Regional and Local)", "Magazines", and "Industry Press" while excluding "Press Releases and Newswires". Unlike the media, which is an independent source of information external to the firm, press releases are firms' self-announcements; thus, they offer different types of information that could affect the interpretation of results (Graf-Vlachy et al., 2020). The list of media outlets obtained has been reviewed to ensure alignment with the categories of interest, and no inconsistencies were identified. We, collected a total of 1,294 news articles related to our sample of startups, published across 93 distinct media outlets (see the full list in Table A1 of the Appendix), with an average article length of 464 words.

Our dataset is structured as an unbalanced panel, with one observation per year per startup, starting from the founding year of the startup up to 2010. Since founding years differ across startups, the number of years (and thus observations) varies by startup.

also, in a narrower sense, of the "existing but unobservable quality of a start-up's technology" (Hoenig & Henkel, 2015). As our paper is related to signals that startups are able to transmit to attract VC attention, we used patent applications (Graham, Stuart J.; Sichelman, Ted, 2008; Hoenen et al., 2014).

### 3.3 Econometric strategy and variables

### 3.3.1 The survival model and the dependent variable

In entrepreneurship research, timing is decisive in determining a startup's survival and competitive advantage (Hallen & Eisenhardt, 2012; McMullen & Dimov, 2013). For startups, the window of opportunity to secure funding can be fleeting, and the time to achieve key milestones, such as securing venture capital, can determine their ability to exit the 'valley of death' and scale up. Media exposure can create an initial spike in attention or interest, thus increasing the probability of obtaining VC funding (e.g., Petkova et al., 2013). However, the effect of media is typically short-lived and quickly fades, which means that startups must capitalize on these brief windows of opportunity by acting swiftly. Adopting a survival framework allows a granular understanding of how quickly media memorability takes effect in securing funding, which is crucially important for startups operating in a fast-moving and competitive environment where every moment counts.

To explore the effect of media memorability on the first VC funding, we use a survival regression (Cox, 1972) estimating the following model:

$$h_i(t) = h_0(t) \cdot \exp(\beta\, \mathbf{X}_i(t) + \gamma\, \mathbf{Z}_i)$$

where $h_0(t)$ is the baseline hazard rate, $\beta$ and $\gamma$ are vectors of coefficients for time-varying covariates X and time-invariant covariates Z, respectively. The model assesses the duration between the startup's foundation year ($\tau_i^0$) and the year of the first round of VC finance ($\tau_i^F$), conditional on not obtaining such finance until that time (Bertoni et al., 2011; Gompers & Lerner, 1999; Nahata, 2008). Once a startup receives the first VC funding, it is no longer at risk of the event "first VC funding" and it is removed from the risk set, meaning it no longer contributes to the likelihood function. Startups that do not receive VC funding by 2010 (the end of the observation period) are treated as right-censored at that point. Accordingly, in order to run our survival analysis, we built the dummy variable *First VC Funding,* which takes the value 1 in the year the startup receives its first VC funding ($\tau_i^F$) and is 0 otherwise (i.e., before and after the event). A positive (negative) coefficient implied a higher (lower) hazard of first VC funding and a shorter (longer) expected duration of the period in the set of firms at risk of VC funding. The set of covariates $X_j(t)$ includes several media characteristics and startup features described in the following sections and reported in Table 1, some of which time-variant and others time-invariant.

### 3.3.2 Independent variable: media memorability

Our primary independent variable captures the concept of startups' media memorability in year *t* using a novel indicator, the Semantic Brand Score. This indicator integrates text mining techniques with social network analysis (Fronzetti Colladon, 2018) to simultaneously consider startup names' prevalence, distinctiveness, and connectivity in news articles. Prevalence refers to how frequently a startup's name appears in news articles during year t. It does not merely count how many media articles mention a startup, as earlier studies on media coverage did (e.g., Petkova et al., 2013); instead, it counts how many times the startup's name appears, including multiple times in the same article. The idea is that the more repeatedly a startup is mentioned, the more easily it is remembered by readers, thereby increasing their awareness of its existence. Unmentioned startups will have zero prevalence in the news, thereby making their recognition and recall more difficult.

To assess the concept of distinctiveness and connectivity included in the composite indicator, it is necessary to consider word relationships within the news text. We achieve this by constructing multiple semantic networks, each corresponding to a specific year of news in our sample, where words/concepts are represented as nodes. Word co-occurrences determine edges connecting nodes, with each edge being weighted by the frequency of co-occurrence between two words. Before building the semantic network, it is necessary to pre-process textual data to ignore lower/upper case, eliminate links and stop-words (such as the words "the" and "and" that usually add little meaning to the text and appear in almost all documents), and remove word affixes to reduce inflectional forms and assign, for example, the word "car" and the word "cars" to the same node – a process known as stemming (Willett, 2006).

Prevalence alone is not sufficient to capture media memorability fully. We could have a frequently mentioned startup (high prevalence), but always in association with the same terms and pieces of information, thus making its image pretty narrow. Accordingly, distinctiveness integrates the information provided by prevalence by looking at the textual associations of a startup's name. It measures the heterogeneity and uniqueness of the textual associations that help build a startup's image. Past research has uncovered the benefits of achieving more numerous and distinctive associations (Grohs et al., 2016). Consistently, distinctiveness is calculated, in each year *t,* through a social network analysis indicator named distinctiveness centrality, while considering the network of word co-occurrences in the text. We use the following formula, taken from Fronzetti Colladon and Naldi (2020), to calculate the distinctiveness of startup name *i*:

$$Distinctiveness\ (i,t) = \sum_{\substack{j=1 \\ j \neq i}}^{g} I\big(w_{ij} > 0\big) \log_{10} \frac{g-1}{d_j}$$

where $g$ is the number of terms in the semantic network, $d_j$ is the degree, i.e., the number of connections, of term j with other words in the network, and $I\big(w_{ij} > 0\big)$ is a function   equal to 1 if there is a connection between terms i and j (that is, if $w_{ij} > 0$, where $w_{ij}$ represents the weight of the edge connecting i and j), and 0 otherwise. In other words, $w_{ij}$ is greater than zero only when an arc exists between the terms i and j. The formula uses a logarithmic term to penalize a startup name's connections to words connected to many other terms in the network (thus not being distinctive associations). Distinctiveness is higher when a startup name is associated with different words in the news and when these associations are unique or very distinctive. For example, startups mentioned for specific characteristics of their products – which are not frequently associated with other new ventures – will have high distinctiveness centrality.

The third dimension, connectivity, measures the brokerage power of a startup's name in the discourse. When considering the network of textual co-occurrences, this factor involves calculating the weighted betweenness centrality (Borgatti et al., 2013; Brandes, 2001) of the startup's name. Indeed, startups with high betweenness centrality connect different words (or topics) in the discourse. An example might be a startup that has developed a new nanotechnology allowing DNA and other materials to be placed inside cells, an activity very useful for research into genetic disorders. In this case, the startup's name could become a connecting bridge between the concepts of DNA, genetic disorders, and nanotechnology. To measure our dimension, we apply the following formula of weighted betweenness centrality for the generic startup name $i$ in each year $t$:

$$Connectivity(i, t) = \sum_{j<k} \frac{p_{jk}(i)}{p_{jk}}$$

where $p_{jk}$ is the number of shortest network paths connecting the terms $j$ and $k$, and $p_{jk}(i)$ is the number of those paths that include the term $i$.

To calculate *Media memorability* of company names in media news for year $t$, we combined prevalence, distinctiveness, and connectivity through a composite indicator to simultaneously account for the interrelations among these three dimensions (Fronzetti Colladon, 2018). Since each year's semantic network was derived from the year's news, resulting in variations in nodes and links across years, and each construct has its range of variation, we standardized the raw scores of distinctiveness and connectivity by considering the scores of all the terms within each year's semantic network. This approach ensures that values across different timeframes are comparable. Because our standardization procedure involved subtracting the mean and dividing by the standard deviation of raw scores, the resulting values can either be positive (if above the mean) or negative (if below the mean)[5]. When a startup is never mentioned in the news, its raw scores for prevalence, distinctiveness, and connectivity are set to zero, as the media cannot impress its name in the memory of stakeholders. Consequently, when a startup is never mentioned in the news, the standardized values of these variables are negative.

### 3.3.3. Moderating variable: media favorability

*Media favorability* is the moderating variable and it is assessed according to Deephouse (2000). We classified media coverage manually (Wei et al., 2017). Two researchers read the full-text versions of all sampled articles separately and coded them into favorable, unfavorable, and neutral (Deephouse, 2000). If there were more than one startup in an article, we coded them independently. Favorable news was classified if a startup is praised for its actions or involved in events that may increase its reputation, for example, affiliation with high-status actors, like Nobel Prize winners, or reception of an award for its technology. An unfavorable record is classified if a startup is criticized for its actions or involved in some events that may decrease its reputation, like Toyota's recalling vehicles because of the problem with accelerator pedals. Neutral news is classified if a startup is reported without any evaluation. Most of the articles were clearly positive, negative, or neutral, and the two coders agreed in 88% of the cases, thus providing inter-coder reliability (Freelon, 2013). We resolved disagreements through discussion, and the final coding was verified by a third researcher. Thus, the variable *Media favorability* was calculated,

---

[5] Prevalence was also standardized using the same approach, with the only difference being that it is based solely on term frequency and does not require the construction of a semantic network.

for startup $i$ in year $t$, as follows:

$$Media\ favorability\ (i,t) = \begin{cases} \dfrac{f^2 - u}{(\text{total})^2} & if\ f > u \\ 0 & if\ f = u \\ \dfrac{f - u^2}{(\text{total})^2} & if\ u > f \end{cases}$$

where "f" is the number of favorable news, "u" is the number of unfavorable news, and "total" is the number of total news. *Media favorability* ranges from -1 to 1, with -1 implying a totally unfavorable reputation, and 1 a totally favorable one.

### 3.3.4 Control variables

In addition to our main variables of interest, we included several control variables. We first controlled for *Media topics* covered by the news. To extract the main discourse themes of startups' textual associations, we carried out a topic modeling of all the news included in our dataset. This procedure involved clustering words by using the Louvain algorithm (Blondel et al., 2008). The Louvain clustering algorithm was applied to the co-occurrence network generated considering all the news in our sample. This algorithm allows the partitioning of the network to identify clusters of words, which represent different discourse topics. Each topic (cluster) is better described by the words that recur frequently, have strong connections within their cluster and weak connections outside. Accordingly, we identified the most representative words of each cluster through a measure of importance that follows this exact logic (Fronzetti Colladon & Vestrelli, 2025). Five topics emerged from the analysis. By design, the topics remain constant over time. For each topic, we created a set of variables that measure the strength of its connection with the different startups, which does vary over time. The topics that emerged from the analysis were labeled according to their most representative keywords. They are briefly described in the following.

*Topic 1 - Public engagement.* This topic refers to the public engagement of startup teams. Such reports often contain interviews with startups' employees or directors and quote their statements. Quoted employees offer a perspective on their business and its impact on people and society.

*Topic 2 – Product & Technology.* News on this topic is related to the startups' existing products, technology and innovation describing their characteristics, manufacturing processes, materials, and applications. Reports cover startups' research, as well as patents and innovation opportunities.

*Topic 3 - Business growth.* This topic is related to the business growth of startups in terms of company value. Words in this topic refer to company shares, sales, profits and their time trends.

*Topic 4 – Management team.* This topic is related to the startups' management teams. Such reports present profiles of startups' directors, discussing their backgrounds and relevant facts about them.

*Topic 5 - Financial market.* The content of this topic is related to the financial market in general, reporting its indexes and trends.

In the final step, we calculated the extent to which the news related to each startup *i*, in each year *t*, covered the different topics. This allowed us to construct a topic intensity measure, defined by the following formula:

$$T_k\ (i,t) = \frac{1}{N} \sum_{w \in D_i^t} I(w \in T_k)$$

here the association value of topic $k$ with startup $i$ in the year $t$ is denoted as $T_k\ (i,t)$. It is calculated as the number of words in the news articles mentioning startup *i* in year *t* that match the keywords of topic *k*, divided by *N*, the number of those news articles. $D_i^t$ represents all words from these articles, and $I(w \in T_k)$ is an indicator function that equals 1 if the word *w* belongs to the set of keywords of topic *k*, and 0 otherwise. The obtained values were standardized before inclusion in the regression models.

Moreover, we included a set of control variables at the startup level, which the extant literature has widely analyzed as determinants of VC financing (Baum & Silverman, 2004; Conti et al., 2013; O'Shea et al., 2008). In particular, the team's human capital is important for further technological upgrading and transferring basic technology into practical technology (Beckman et al., 2007; D. H. Hsu, 2007), while scientific and technological skills give startups a greater opportunity to survive (Giarratana & Torrisi, 2010). We, thus, controlled for *Team heterogeneity* through a time-varying Herfindahl-Hirschman Index calculated in each year $t$ based on the distribution of directors' professional orientation (Hambrick et al., 1996) across six occupational background categories: accounting, general management, law, engineering, finance, and other. A management team with diversified occupational backgrounds has a greater ability to identify and solve a more comprehensive set of problems a startup has to deal with (Anderson et al., 2011; Bantel & Jackson, 1989). Greater occupational background diversity allows for a better division of labor and the exploitation of complementarities and cross-fertilization, which are good for creativity and productivity. In each year $t$, *Academic prevalence* measures the proportion of directors on the team with an academic background. As technology commercialization is linked to the development of the underlying science (Munari & Toschi, 2011; Pisano, 2006; Zucker et al., 1998), external investors recognize the high quality of directors with an academic background in the entrepreneurial team. Accordingly, we also included the number of publications from the startup's directors (*Total publication*), as well as the startup's stock of patents filed to the European Patent Office (EPO) (*Patents*) at the time of the first VC funding ($T_i^F$) or up to 2010 in case of no reception of VC funding. We also included a time-invariant dummy variable (*Academic spinoff*) that took a value of 1 if a startup is an academic spinoff (ASO). Previous literature shows that ASOs are more distant from the business environment and therefore face more difficulties in attracting VC funding (Munari & Toschi, 2011). On the other hand, the strong scientific capabilities of ASOs increase the possibility of developing high-quality technology, which is a critical factor that VCs consider. Finally, the time-invariant dummy *London* helps to control for the presence of a VC hub in the area.

The operationalization of all the variables used in our econometric models is summarized in

Table 1.

**Table 1**. Description of variables (overview)

| VARIABLE | DEFINITION | SOURCE |
|---|---|---|
| First VC funding | A dummy variable that takes the value of 1 in the year a startup (*i*) first receives venture capital funding (denoted as $T_i^F$), and 0 in all other years within the observation period ending in December 2010 (*T*). | Thomson One |
| Age | Difference between the year of the first VC funding ($T_i^F$) and the founding year of company i ($T_i^0$) | |
| Media memorability | The memorability of startups' names in media news in year *t*, measured by prevalence, distinctiveness, and connectivity. | LexisNexis |
| Prevalence | The number of times a startup's name appears in news articles of year *t*. | LexisNexis |
| Distinctiveness | $\sum_{\substack{j=1 \\ j\neq i}}^{g} I(w_{ij} > 0) \log_{10} \frac{g-1}{d_j}$ <br> Calculated, in each year *t*, based on the semantic network constructed from the news of year *t*. | LexisNexis |
| Connectivity | $\sum_{j<k} \frac{p_{jk}(i)}{p_{jk}}$ <br> Calculated, in each year *t*, based on the semantic network constructed from the news of year *t*. | LexisNexis |
| Media favorability | $\begin{cases} \dfrac{f^2 - u}{(total)^2} & \text{if } f > u \\ 0 & \text{if } f = u \\ \dfrac{f - u^2}{(total)^2} & \text{if } u > f \end{cases}$ <br> Calculated, in each year *t*, based on news articles that mention the startup in year *t*. | LexisNexis |
| Team heterogeneity | A time-varying Herfindahl-Hirschman Index calculated, in each year *t*, based on the distribution of directors across six occupational background categories: accounting, general management, law, engineering, finance, and other. | Company Director Check FAME |
| Academic prevalence | A time-varying ratio calculated, in each year *t*, as the number of directors holding PhD degrees divided by the total number of directors on the board. | FAME |
| Patents | The logarithm of 1 plus the patent stock of the startup, measured up to $T_i^F$ for VC-backed startups or up to *T* for startups that have not received VC funding. | QPAT |
| Publications | The total number of publications authored by all of a startup's directors up to $T_i^F$ for VC-backed startups, or up to *T* for startups that have not received VC funding. | Scopus |
| Academic spin-off | A time-invariant dummy variable equal to 1 if the focal startup is classified as an academic spinoff, and 0 otherwise. | MNT |
| London | A time-invariant dummy variable equal to 1 if the startup is located in London, and 0 otherwise. | MNT |
| Topic 1 – 5 | A set of five time-varying variables measured annually, indicating the strength of a startup's association with specific discourse topics. The topics are numbered from 1 to 5, corresponding respectively to: (1) public engagement, (2) product and technology, (3) business growth, (4) management team, and (5) financial market. | LexisNexis |

## 4 Results

### 4.1 Descriptive analyses

Table 2 presents the descriptive statistics for the main variables for the full sample of 197 startups and the split sample of VC-funded and non-VC-funded startups. Overall, 138 out of 197 startups (70%) have been reported at least once in the media. Among the VC-backed startups, 57 out of 66 startups (86.36%) have been reported at least once, while among the non-VC-backed startups, 81 out of 131 (61.83%) have been reported. For the split sample, a t-test to assess differences between the two groups is also provided, showing significant differences along the different dimensions. In particular, the value of *Media memorability* for VC-backed and non-VC-backed is statistically different at the 0.05% level. Also, the mean value of *Media favorability* is positive for both groups. This result indicates that reports conveying a positive message are prevalent among news concerning the whole sample of technology-based startups. However, the t-test also indicates that the average of media favorability is statistically different between the two groups at the 1% level, being more positive for VC-backed startups.

Table 3 reports the correlation matrix of the main variables. All the coefficients are small, except for *Prevalence*, *Distinctiveness,* and *Connectivity*, which are strongly correlated to each other, which motivates using a composite indicator of media memorability that considers the three dimensions simultaneously.

**Table 2.** Descriptive statistics and t-tests

| Variable | Full Sample (197 startups) | | | | VC-funded (66 startups) | | | | Non VC-funded (131 startups) | | | | Diff. |
|---|---|---|---|---|---|---|---|---|---|---|---|---|---|
| | Mean | Std. Dev. | Min | Max | Mean | Std. Dev. | Min | Max | Mean | Std. Dev. | Min | Max | |
| First VC funding | 0.335 | 0.473 | 0 | 1 | 1 | 0 | 1 | 1 | 0 | 0 | 0 | 0 | |
| Age | 7.434 | 4.142 | .1 | 15 | 2.598 | 2.342 | 0 | 10 | 9.870 | 2.295 | 6 | 15 | *** |
| Prevalence | -0.346 | 0.202 | -0.848 | 1.341 | -0.250 | 0.316 | -0.848 | 1.341 | -0.394 | 0.068 | -0.407 | 0.202 | *** |
| Distinctiveness | -0.087 | 0.020 | -0.201 | -0.054 | -0.072 | 0.029 | -0.201 | -0.054 | -0.095 | 0 | -0.095 | -0.094 | *** |
| Connectivity | -0.674 | 0.457 | -1.484 | 3.196 | -0.446 | 0.704 | -1.484 | 3.196 | -0.788 | 0.163 | -0.831 | 0.443 | *** |
| Media memorability | 0.159 | 0.351 | 0 | 1 | 0.357 | 0.457 | 0 | 1 | 0.060 | 0.226 | 0 | 1 | *** |
| Media favorability | 0.076 | 0.266 | 0 | 1 | 0.121 | 0.329 | 0 | 1 | 0.053 | 0.226 | 0 | 1 | *** |
| London | 0.361 | 0.230 | 0 | 0.735 | 0.442 | 0.192 | 0 | 0.735 | 0.321 | 0.237 | 0 | 0.722 | * |
| Team heterogeneity | 0.322 | 0.295 | 0 | 1 | 0.342 | 0.246 | 0 | 1 | 0.312 | 0.317 | 0 | 1 | *** |
| Academic prevalence | 0.126 | 0.347 | 0 | 1.792 | 0.333 | 0.518 | 0 | 1.792 | 0.021 | 0.120 | 0 | 0.693 | ns |
| Patents | 0.451 | 0.645 | 0 | 2.262 | 0.463 | 0.616 | 0 | 1.833 | 0.446 | 0.661 | 0 | 2.262 | *** |
| Publications | 0.518 | 0.501 | 0 | 1 | 0.621 | 0.489 | 0 | 1 | 0.466 | 0.501 | 0 | 1 | ns |
| Academic Spinoff | 6.753 | 16.871 | 0 | 115 | 9.591 | 13.011 | 0 | 51.889 | 5.323 | 18.394 | 0 | 115 | ** |
| Topic 1 – Public engagement | 6.258 | 14.627 | 0 | 92 | 9.483 | 13.084 | 0 | 68.500 | 4.632 | 15.133 | 0 | 92 | * |
| Topic 2 – Product & Technology | 3.623 | 9.113 | 0 | 81.5 | 6.074 | 9.760 | 0 | 50 | 2.393 | 8.544 | 0 | 81.500 | ** |
| Topic 3 – Business growth | 3.514 | 8.190 | 0 | 59 | 5.738 | 7.970 | 0 | 50 | 2.388 | 8.098 | 0 | 59 | ** |
| Topic 4 – Management team | 1.221 | 4.326 | 0 | 49.667 | 1.562 | 2.252 | 0 | 7.818 | 1.049 | 5.059 | 0 | 49.667 | ** |
| Topic 5 - Financial market | 0.335 | 0.473 | 0 | 1 | 1 | 0 | 1 | 1 | 0 | 0 | 0 | 0 | ns |

**Table 3.** Correlation matrix

| Variables | (1) | (2) | (3) | (4) | (5) | (6) | (7) | (8) | (9) | (10) | (11) | (12) | (13) | (14) | (15) | (16) | (17) | (18) |
|---|---|---|---|---|---|---|---|---|---|---|---|---|---|---|---|---|---|---|
| (1) First VC Funding | 1.000 | | | | | | | | | | | | | | | | | |
| (2) Age | -0.831* | 1.000 | | | | | | | | | | | | | | | | |
| (3) Prevalence | 0.324* | -0.238* | 1.000 | | | | | | | | | | | | | | | |
| (4) Distinctiveness | 0.337* | -0.270* | 0.932* | 1.000 | | | | | | | | | | | | | | |
| (5) Connectivity | 0.549* | -0.449* | 0.182* | 0.383* | 1.000 | | | | | | | | | | | | | |
| (6) Media memorability | 0.355* | -0.273* | 0.982* | 0.982* | 0.315* | 1.000 | | | | | | | | | | | | |
| (7) Media favorability | 0.401* | -0.237* | 0.519* | 0.430* | 0.217* | 0.491* | 1.000 | | | | | | | | | | | |
| (8) London | 0.121 | -0.021 | -0.012 | -0.032 | 0.020 | -0.020 | 0.033 | 1.000 | | | | | | | | | | |
| (9) Team heterogeneity | 0.251* | -0.267* | 0.060 | 0.027 | 0.116 | 0.050 | 0.044 | 0.041 | 1.000 | | | | | | | | | |
| (10) Academic prevalence | 0.048 | -0.048 | -0.003 | 0.005 | 0.089 | 0.005 | 0.068 | 0.091 | 0.055 | 1.000 | | | | | | | | |
| (11) Patents | 0.426* | -0.388* | 0.096 | 0.142* | 0.408* | 0.152* | 0.207* | -0.005 | 0.152* | 0.055 | 1.000 | | | | | | | |
| (12) Publications | 0.012 | 0.087 | 0.082 | 0.078 | 0.062 | 0.083 | 0.192* | 0.088 | 0.033 | 0.219* | 0.123 | 1.000 | | | | | | |
| (13) Academic spinoff | 0.147* | -0.137 | 0.173* | 0.154* | 0.070 | 0.168* | 0.152* | 0.162* | 0.087 | 0.351* | 0.056 | 0.177* | 1.000 | | | | | |
| (14) Topic 1 - Public engagement | 0.120 | -0.072 | 0.350* | 0.282* | 0.116 | 0.326* | 0.359* | -0.038 | 0.093 | 0.048 | 0.135 | 0.108 | 0.093 | 1.000 | | | | |
| (15) Topic 2 - Product & Technology | 0.157* | -0.087 | 0.397* | 0.376* | 0.137 | 0.395* | 0.384* | -0.035 | 0.078 | 0.043 | 0.111 | 0.108 | 0.071 | 0.871* | 1.000 | | | |
| (16) Topic 3 - Business growth | 0.191* | -0.137 | 0.373* | 0.338* | 0.157* | 0.366* | 0.400* | -0.005 | 0.033 | 0.030 | 0.204* | 0.068 | 0.111 | 0.821* | 0.829* | 1.000 | | |
| (17) Topic 4 – Management team | 0.193* | -0.095 | 0.365* | 0.323* | 0.180* | 0.356* | 0.371* | -0.002 | 0.070 | 0.039 | 0.220* | 0.139 | 0.112 | 0.853* | 0.827* | 0.841* | 1.000 | |
| (18) Topic 5 - Financial market | 0.056 | 0.001 | 0.201* | 0.146* | 0.070 | 0.181* | 0.213* | -0.024 | -0.051 | -0.028 | 0.089 | 0.056 | 0.126 | 0.572* | 0.576* | 0.582* | 0.549* | 1.000 |

*** p<0.01, ** p<0.05, * p<0.1

**4.2 Regression analyses**

Table 4 presents the hazard ratios (HRs) of our survival analyses. Model 1 includes only the controls and media favorability. Models from 2 to 4 assess the effect of the three components of media memorability introduced separately (it is not possible to use them in the same model due to collinearity problems, as shown in Table 3) and test for the effect of each dimension of media memorability in isolation. Model 5 considers the presence of our composite indicator of media memorability as the main effect, while Models 6 to 9 explore the interactions of media memorability and its components with *Media favorability*. We checked for potential multicollinearity problems in the full model (9) and found no evidence of it in our setting, as the Variance Inflated Factor shows an average value equal to 3.34.

**Table 4.** Survival analysis for the likelihood of receiving VC funding (sample of 197 startups)

| VARIABLES | Hazard Ratios | | | | | | | | |
|---|---|---|---|---|---|---|---|---|---|
| | (1) | (2) | (3) | (4) | (5) | (6) | (7) | (8) | (9) |
| London | 1.130 | 1.140 | 1.134 | 1.127 | 1.137 | 1.141 | 1.134 | 1.126 | 1.137 |
| | (0.120) | (0.122) | (0.121) | (0.120) | (0.121) | (0.122) | (0.121) | (0.120) | (0.121) |
| Team heterogeneity | 1.567** | 1.595*** | 1.614*** | 1.565*** | 1.604*** | 1.578*** | 1.613*** | 1.543*** | 1.596*** |
| | (0.237) | (0.240) | (0.243) | (0.237) | (0.241) | (0.238) | (0.243) | (0.234) | (0.241) |
| Academic prevalence | 0.991 | 1.050 | 1.046 | 0.989 | 1.051 | 1.054 | 1.047 | 0.975 | 1.055 |
| | (0.143) | (0.153) | (0.153) | (0.143) | (0.154) | (0.153) | (0.153) | (0.142) | (0.154) |
| Patents | 1.324** | 1.347*** | 1.333*** | 1.309*** | 1.340*** | 1.348*** | 1.333*** | 1.319*** | 1.339*** |
| | (0.130) | (0.136) | (0.134) | (0.130) | (0.136) | (0.136) | (0.134) | (0.131) | (0.135) |
| Publications | 1.283* | 1.270 | 1.253 | 1.283* | 1.262 | 1.276 | 1.254 | 1.297* | 1.266 |
| | (0.191) | (0.191) | (0.188) | (0.192) | (0.190) | (0.192) | (0.189) | (0.196) | (0.191) |
| Academic spinoff | 1.113 | 1.092 | 1.083 | 1.117 | 1.088 | 1.074 | 1.080 | 1.114 | 1.078 |
| | (0.152) | (0.152) | (0.150) | (0.153) | (0.151) | (0.152) | (0.152) | (0.153) | (0.152) |
| Topic 1 - Public engagement | 0.654 | 0.580 | 0.616 | 0.652 | 0.591 | 0.586 | 0.617 | 0.654 | 0.595 |
| | (0.230) | (0.234) | (0.242) | (0.230) | (0.237) | (0.234) | (0.243) | (0.231) | (0.237) |
| Topic 2 – Product & Technology | 1.441 | 1.307 | 1.288 | 1.417 | 1.292 | 1.314 | 1.289 | 1.434 | 1.295 |
| | (0.426) | (0.417) | (0.403) | (0.422) | (0.410) | (0.416) | (0.403) | (0.426) | (0.410) |
| Topic 3 - Business growth | 1.252 | 1.256 | 1.265 | 1.256 | 1.270 | 1.270 | 1.266 | 1.227 | 1.276 |
| | (0.383) | (0.422) | (0.420) | (0.382) | (0.425) | (0.425) | (0.420) | (0.378) | (0.426) |
| Topic 4 – Management team | 0.688 | 0.616 | 0.621 | 0.686 | 0.613 | 0.646 | 0.625 | 0.659 | 0.632 |
| | (0.575) | (0.555) | (0.555) | (0.568) | (0.552) | (0.575) | (0.559) | (0.550) | (0.565) |
| Topic 5 - Financial market | 1.480 | 1.599 | 1.576 | 1.482 | 1.589 | 1.598 | 1.578 | 1.503 | 1.590 |
| | (0.484) | (0.560) | (0.547) | (0.484) | (0.555) | (0.559) | (0.547) | (0.493) | (0.555) |
| Media favorability | 1.480*** | 1.430*** | 1.433*** | 1.477*** | 1.431*** | 1.405*** | 1.429*** | 1.499*** | 1.415*** |
| | (0.163) | (0.163) | (0.165) | (0.162) | (0.164) | (0.168) | (0.169) | (0.165) | (0.168) |
| Prevalence | | 1.642*** | | | | 1.493* | | | |
| | | (0.195) | | | | (0.318) | | | |
| Distinctiveness | | | 1.571*** | | | | 1.544** | | |
| | | | (0.161) | | | | (0.296) | | |
| Connectivity | | | | 1.209 | | | | 1.283 | |
| | | | | (0.229) | | | | (0.272) | |
| Media memorability | | | | | 1.667*** | | | | 1.573** |
| | | | | | (0.194) | | | | (0.316) |
| Prevalence*Favorability | | | | | | 1.056 | | | |
| | | | | | | (0.098) | | | |
| Distinctiveness*Favorability | | | | | | | 1.009 | | |
| | | | | | | | (0.086) | | |
| Connectivity*Favorability | | | | | | | | 0.866 | |
| | | | | | | | | (0.122) | |
| Media memorability*Favorability | | | | | | | | | 1.033 |
| | | | | | | | | | (0.092) |
| AIC | 641.030 | 630.690 | 628.832 | 641.906 | 629.203 | 632.316 | 630.820 | 642.971 | 631.062 |
| BIC | 706.004 | 701.079 | 699.221 | 712.296 | 699.593 | 708.120 | 706.624 | 718.775 | 706.866 |
| LR chi2 | 57.54*** | 69.88*** | 71.74*** | 58.67*** | 71.37*** | 70.26*** | 71.75*** | 59.60*** | 71.51*** |
| Wald chi2 | 90.63*** | 145.72*** | 140.33*** | 92.69*** | 144.32*** | 159.14*** | 150.90*** | 88.33*** | 154.08*** |

*Standard errors are in parentheses  *** p<.01, ** p<.05, * p<.1*

Starting with Model 1, our results show that being located in the London area, having higher occupational heterogeneity among directors, and having more patents positively impact VC investments ($p<0.1$, $p<0.01$, and $p<0.05$, respectively). As expected, media favorability positively and significantly affects VC funding ($p<0.001$).

Models 2-4 show that among the three components of media memorability, *Prevalence* and *Distinctiveness* are significant ($p<0.001$), highlighting the relevant role of textual associations that build a differentiating startup's image. Model 5 tests for Hypothesis 1 and introduces the variable *Media memorability*, which measures the combined effect of prevalence, distinctiveness, and connectivity. The results suggest a positive and significant effect of this dimension ($p<0.001$), thus supporting our H1. The hazard rates show that a 1% increase in media memorability results in a 67% increase in the likelihood of receiving VC funding.

We test Hypothesis 2 in Model 9 by interacting *Media memorability* with the variable *Media favorability*. The results indicate a positive value for the interaction term, but, it is not statistically significant. This suggests that the effect of *Media memorability* on VC funding, combined with a positive general evaluation of the startup, is not higher than that with less favorable tones used in the news. This result does not confirm our second hypothesis. It is worth noting that the interactions of media favorability with the three dimensions of media memorability separately are also not significant (Models 6-8). The insignificant interactions between media memorability and media favorability show that, contrary to our expectations, investors value the two dimensions of media coverage independently. Thus, in making their decisions, VCs are affected by media memorability (and its components) regardless of whether or not the tone of the news is favorable. To the best of our knowledge, this is the first study to explore the interaction of media tone with other dimensions of media, with the partial exception of Wei et al (2017) who analyzed the effect of the interaction between generalized media favorability and a proxy for media visibility (i.e., the number of posts concerning a firm in online stock forums) on a firm's cumulative abnormal adjusted return (CAR). They focused on firms facing a corporate crisis and found that the buffering effect of favorability was weakened by media visibility. While the setting of Wei et al. (2017) differs from ours, their findings suggest that the interaction between media memorability and media favorability may have nuanced effects on VCs' decisions, which deserves further exploration.

### 4.3 Robustness checks

We conducted a series of robustness checks to further corroborate our results. First, to exclude the possibility that our results are mainly driven by a few ventures with very high value of media memorability—that is, a few "star" startups that might have attracted both significant media attention and VC funding—we removed the observations with values of media memorability above three standard deviations from the mean. This corresponds to eliminating 4 observations from our sample. Second, we estimated a logistic regression with standard errors clustered at the startup level, including the usual set of control variables (see Table A2). Third, we compared the impact of media visibility, as measured by the number of news articles a startup receives, with the effect of media memorability. Our analysis revealed that the mere count of news articles citing a startup has a lower impact on VC funding (Hazard Ratios: 1.44, $p<0.05$) compared to the combined influence of the three components of media memorability (Hazard Ratios: 1.67, $p<0.001$). Finally, we run a model to test if the effect of media memorability persists after controlling for media coverage (i.e., number of news). Also in this case, we find that media memorability is still significant (Hazard Ratios: 1.56, $p<0.001$). These findings suggest that our measure provides a valuable additional perspective for understanding the venture capital decision-making process.

## 5 Discussion

In evaluating startup potential, VCs rely on observable attributes that are presumably correlated with additional, unobserved determinants of startups' quality (Stuart et al., 1999). Past research has also investigated the importance of media reputation for startup performance and access to VC funding (e.g., Petkova, 2014; Petkova et al., 2013; Pollock et al., 2008), but it has overlooked the peculiar characteristics of VCs which make this category of stakeholders more able to use media reputation to grasp fine-grained valuable information for their decision-making compared with other stakeholder audiences. Our work recognizes the importance of the VCs' peculiar characteristics and contributes to the research on the VC funding of startups by digging into the dimensions of media reputation affecting VC investment decisions that have not been investigated in previous works.

A venture capitalist we interviewed noted that while VCs are not necessarily aware of measures of media memorability like distinctiveness and connectivity, these "measures of media memorability capture important unobservable dimensions of the entrepreneur/startup capacity to generate value". While VCs can directly assess a startup's technological value and market potential, media allows for signaling "how good an entrepreneur is at communicating and possibly selling, the business idea even in a deep-tech industry". Another VC manager asserted that they "learn a lot from media" and "the dimensions of media memorability resonate with me a lot because they are exactly what we try to convey, perhaps unconsciously, when we talk about communication. We are investing in a startup trying to create a rich image linking its brand to several concepts, which seems to work a lot as investors are calling us to know more about that company. If startups can be in the press with great messages, they signal they are able at what they do".

### 5.1 Theoretical contribution

Our research provides several implications for scholarship on the role of media reputation in startup financing and VC decision-making. Our findings complement and extend the academic scholarly conversation on the following grounds.

First, we contribute to the literature on the role of media reputation in startups' VC funding by introducing a novel construct, *media memorability*, which builds on the notions of brand knowledge from the marketing communication literature (Keller, 1993), associative network memory model from cognitive psychology (Collins & Loftus, 1975; Keller, 1993; Krishnan, 1996), and connectivity (a measure of network centrality) from social network analysis (Burt, 2004).

Our exploration of media memorability in the entrepreneurial finance setting allows us to dive deeper into the dimensions of media coverage that make a startup more distinctive and attractive to VCs. While the importance of these dimensions is recognized by brand equity scholars in marketing communication and cognitive psychology, entrepreneurial financing research has not tried to investigate the mechanisms through which media coverage contributes to generating distinctiveness and connectivity of a startup name in the memory of VC investors. Our findings extend entrepreneurial financing research by arguing and showing that media contribute to startup financing by signaling a startup's distinctiveness and connectivity. This suggests that media constitute an important source of finer-grained information about the quality of a startup.

Second, accounting for the importance of media memorability is also useful to scholars interested in new organizations' legitimacy as a means to ease access to key resources such as finance (Petkova et al., 2013), human resources (Williamson, 2000), and partners (Pollock & Gulati, 2007). Media are information intermediaries that function as a filter for new developments and new organizations. Thus, media constitute an institution that legitimizes new organizations by signaling to stakeholders their quality (Petkova et al., 2013). Earlier studies have insisted on the role of media as informed intermediaries who evaluate and select innovative entrepreneurial projects, thus providing legitimacy advantages to new firms and moderating the information asymmetry that characterizes the VC funding of startups (e.g., Audretsch et al., 2009; Rindova et al., 2007). The literature on media as a mechanism of legitimation of new organizations, however, provides limited insights into how the position of a new organization in the media attracts more market coverage and affects its perceived value in the eyes of sophisticated stakeholders like VCs. The main contribution of our study to current research on media legitimation of new organizations is that, beyond the legitimation effect produced by the mere frequency and tone of a startup's citations in the news, media allows VCs to dive deeper into key attributes of a startup such as distinctiveness and connectivity that can be crucially important in their search for fine-grained information about the quality and market potential of startups.

In a world where technology startups are particularly exposed to (and solicit) media coverage, the significant amount of information circulating through the media helps VCs better understand the potential value of startups. We think that analyzing media reputation through the notion of media memorability provides a novel and valuable contribution to the literature on the role of media in new venture funding, which has focused on media coverage and favorability without considering the importance of the distinctiveness and the connectivity of a brand whose effects on VC decision making can be understood from the analysis of characteristics of media news. Future research will benefit from our study to further explore media as a legitimating institution for startups in other empirical settings. Moreover, future research could explore whether the legitimation effect of media memorability varies with boundary conditions such as the level of a startup's innovativeness and the type of investors. Recently, Mochkabadi et al.

(2024) have found that media functions as a legitimacy amplifier for incrementally innovative firms, while it does not significantly moderate the negative legitimation effect of radically innovative firms in equity crowdfunding. Future studies could extend this line of research to see whether the moderating effects of media coverage observed in equity crowdfunding also hold for media memorability of innovative startups in search of VC financing.

Third, we contribute to a deeper understanding of VC decision-making. VCs are expert, informed stakeholders who typically rely on a precise checklist of criteria to support their investment decisions (e.g., entrepreneurial team, product/service attractiveness, technology, market, and competitive conditions). However, studies from cognitive psychology found that VCs lack introspection capacity, which may limit their understanding of the decision-making process and learning (Shepherd & Zacharakis, 2002). Moreover, because of the high technological and market uncertainty characterizing nanotechnology, a large share of VC-backed ventures fail to provide returns to investors, supporting the view that VC decision-making takes place under conditions of bounded rationality (Simon, 1991). Previous research has shown that VCs' decisions are affected by the media, even though VCs are a well-informed audience with direct access to firm-specific information (Petkova et al., 2013). We contribute to this line of research by showing that VCs rely on (or behave as if they were aware of the value of) information provided by the characteristics of media coverage. In particular, the effect of distinctiveness and connectivity on VC financing suggests that media not only select and filter information but also signal the perceived value of a startup in the eyes of other audiences, such as customers or partners. Greater awareness and understanding of media news content, a still unexplored research area, thus could further moderate VCs' cognitive biases and improve their decision-making. We extend this line of research by arguing that a deeper exploration and more effective exploitation of information provided by the characteristics of news will benefit from a 'scientific approach' to decision-making (Briner et al., 2009; Camuffo et al., 2020; Felin & Zenger, 2015) based on reasonable facts rather than on first impressions (Hassan & Hanapi, 2013). Future research on VC decision-making could identify the actual adoption of a scientific approach and the specific mechanisms through which a 'scientific' use of characteristics of news affects VCs' ability to select and gain from their investments in startups.

## 5.2 Implications for business practice

To survive and prosper in the early stages of their life, startups must attract the attention of providers of financial resources. Earlier work has found that the interest of specialized industry media stimulated by startups' sense-giving strategy is positively associated with the level of VC funding obtained by startups (Petkova et al., 2013). Extant research has also demonstrated that the intensity or frequency of sense-giving activities increases brand awareness by providing more information or making information more readily available (Kennedy, 2008; Pollock et al., 2008). Various types of sense-giving activities have proven effective in attracting media attention, such as involvement in interactive events – such as conferences, trade shows, professional gatherings, technology contests, and other industry events (Petkova et al., 2013). For instance, technology start-ups that frequently participate in industry conferences are more likely to be noticed and enjoy reputational benefits (Fleming & Waguespack, 2007). New firms could also use activities like website updates and knowledge dissemination instruments like white papers, books, and presentations to increase their visibility. Moreover, combining different types of communications in their press releases enables new organizations to convey richer, more comprehensive information (Petkova et al., 2013). Using a wide set of communication channels, startups can exploit the complementarity between traditional media channels and social media to circulate engaging narratives and highlight key assets such as the management team, founders' experience, innovations, and partnerships. Media can also be solicited to communicate endorsements from credible industry experts, showcase the startup's role in creating jobs, or spread information that enhances a startup's positive image. Startups could improve the equity value of their brand in the eyes of VCs by marketing activities fostering brand associations in the media to arouse feelings, thoughts, and experiences that can influence consumers' responses and purchase intentions (Grohs et al., 2016; Keller, 2003), make their brand name more embedded in the media's discourse and strengthen its memorability. Other potential activities to strengthen recall and associations include advertising on investor-focused channels, industry platforms, and forums, collaboration with journalists and influencers to generate feature stories or interviews, and engagement with events, webinars, or behind-the-scenes tours to increase personal connections.

Our findings suggest that startups can influence venture capital funding when their sense-giving efforts in increasing media coverage succeed in embedding their names in the memory of venture capitalists. The importance of distinctiveness and connectivity for VC funding revealed by our analysis, in particular, could inform and help startup founders and top managers direct their attention to sense-giving activities that are more likely to solicit an evaluative or behavioral response, impacting memorability of the company's brand value and attracting VCs' attention. Our findings suggest that besides media coverage, how the startup's story is told may affect VC funding. This complements the entrepreneurship literature that suggests that narratives and storytelling approaches implemented by entrepreneurs serve as powerful tools for embedding uniqueness and distinctiveness in identity construction (Taeuscher et al., 2021). Accordingly, startups should prioritize distinctive, memorable messaging highlighting uniqueness, clarity, and impact. Related to this, entrepreneurs can explicitly invest in narrative building by engaging with journalists to craft compelling stories that emotionally or cognitively resonate with readers. From this perspective, startups need to be aware of the different effects of narratives on different types of investors. Studies on crowdfunding investment suggest that image-based rhetoric, which refers to language that readily stimulates an intense sensory experience, such as a mental picture or sound, induces a stronger emotional response and improves the persuasiveness of the narrative, particularly for novice crowdfunding investors. Instead, concept-based rhetoric, which is logical, grounded in reality, and associated with systematic, rational thinking, has a more substantial appeal for potential investors with more significant experience in crowdfunding investments (Patel et al., 2021). As VC investors are expert and sophisticated decision-makers, actions to increase media memorability should draw on a concept-based rhetoric that, through the media, could prime VC investors to examine the information received. Other narrative dimensions are also likely to be effective in attracting financial resources for startups. For example, resourcefulness narratives are discursive accounts of entrepreneurial actions that describe entrepreneurs as using, assembling, or deploying resources creatively to overcome an impediment. Resourcefulness narratives generate positive emotional and cognitive reactions from external resource providers, which may lead them to support a new venture (Fisher et al., 2021). Moreover, given the short-lived nature of media effects, startups should consider the key role of timing and align media exposure with funding windows, ensuring that memorable stories are circulating at the time when VCs are actively searching for deals.

Finally, from the investors' point of view, the concept of media memorability can be seen as a screening cue. Memorable media coverage can be a signal of quality communication, public interest, or media-savvy founders, which can complement other due diligence factors (e.g., team, tech, market). Moreover, our research highlights the presence of potential bias in VC deal flow as VCs might unconsciously favor startups with more memorable media exposure, potentially overlooking promising but less-visible ventures. Being aware of this cognitive mechanism and bias can help investors reflect on why certain startups seem more salient and can promote more equitable and diversified investment decisions.

### 5.3 Limitations and future research

This study has limitations that open new inquiries for future research. First, we focus on VCs as the only funding source for startups. The effect of media memorability on VCs could not be generalized to other types of resource providers; therefore, a comparison among different actors could be valuable in future research. Indeed, the literature on entrepreneurial finance suggests a broad spectrum of investors that could be differently influenced by media memorability as they adopt different approaches to startup evaluation. For instance, crowdfunding investors can select projects from a broad pool of options available on a crowdfunding platform. In these settings, investors prefer to support projects of entrepreneurs they know personally or professionally or with a consolidated media reputation (Aleksina et al., 2019; Borst et al., 2018; Scheaf et al., 2018). The impact of media memorability may also be higher for public VC investors compared to private VCs, as they are generally interested not only in financial goals but also in social objectives, with strong attention to local development, and possess lower skills and competencies in evaluating and supporting technology-based startups (Leleux & Surlemont, 2003; Lerner, 1999; Munari & Toschi, 2015). Second, our analysis does not account for the power of media memorability in the minds of other stakeholders, such as customers or partners. These indirect effects could influence VC investment decisions as greater media memorability implies improved perceptions of product performance, greater customer loyalty, less vulnerability to competitive marketing actions, and larger margins (Hoeffler & Keller, 2003; Keller, 1993). Thus, the visibility of a startup in the media discourse and the level of collective attention by a vast audience are likely to make VCs more confident about the success of their investments. While our data do not allow us to test for these mechanisms, future research could investigate how the effect of media memorability on relevant audiences, like a startup's customers, influences VC investment decisions.

Third, we focus only on traditional media as a source of information for institutional investors. As discussed in the description of the context of this research, we selected the period 1995-2004 as it is the incubation phase of the nanotech industry, characterized by high degrees of technological, economic, and regulatory uncertainty. In this phase of the industry life cycle, traditional media were key actors in tracing the emergence of the industry and providing relevant information for stakeholders. In this period, the use of social media was still not influential. The dawn of social media started in the 2000s with the introduction of LinkedIn in 2003, Facebook in 2004, YouTube in 2005, and Twitter in 2006, but their diffusion and impact have grown later on due to technological advancements like the launch of the iPhone in 2007. We are aware that over the years, a new field generally becomes more institutionalized, uncertainty decreases, and the media may become less useful for VC evaluations. Although various indicators suggest that the nanotechnology field is still far from its maturity stage, future research could focus on more recent stages of this industry and account for potential changes in the effect of media in a world where the use of online communication has become widespread and VCs are more informed about nanotechnology.

While social media probably cannot offer the same degree of legitimation as professional media, we recognize the limitations of excluding social media from our analysis. Incorporating digital media data, such as social media platforms, blogs, and online news sources, would, for instance, provide a more comprehensive understanding of how media visibility influences VC funding decisions along different channels. Digital media has become an increasingly significant source of information and engagement for entrepreneurs and investors, often shaping perceptions and opportunities in real-time. Thus, future research could leverage data analytics and social media metrics to assess how narratives reported in social media affect the speed and likelihood of securing VC, and whether the effect differs significantly from that produced by professional media. This broader perspective would provide future research with actionable insights into the evolving role of media in VC dynamics.

The fourth limitation concerns endogeneity. Our data suggest caution about the causal interpretation of the linkages between the main variables in our framework. Still, our findings about the association between our explanatory variables and VC funding suggest that future research could benefit from the richness of information offered by a deeper examination of the characteristics of news to explore the mechanisms through which this source of information affects VCs' evaluation and decision-making.

Another limitation originates from the sample of startups, which are all operating in the UK and a single sector. Future research could replicate our study for different countries and business sectors to assess, for example, the impact of media memorability on VC funding for startups operating in new fields that receive more media coverage and are characterized by high technological and market uncertainty, such as Generative AI search tools for which an organized market is still in a nascent stage and various standards are competing in the market such as ChatGPT (OpenAI and Microsoft), Cloud (Anthopic) and Apprentice Bard (Google/Alphabet) (Ferràs-Hernández et al., 2023; Mariani & Dwivedi, 2024).

Moreover, our study assumes that all the startups in our sample sought VC funding. We believe that VC as a funding source is equally important for most firms in our sector, as is the case with other industries characterized by rapid technical change and high R&D and commercialization costs. Thus, biases arising from unobserved heterogeneity and self-selection are probably unimportant in our empirical setting. However, it may be likely that some companies in our sample were not actively seeking VC funding. Our results should be interpreted with this limitation in mind, and future studies should explore this topic further.

Lastly, we focused on the impact of media memorability on the VC decision to provide the first round of investments. Our data do not allow us to establish whether and how the effect of media memorability varies along the different stages of the VC investment process. However, investors' interest in the information provided by media could be different during scouting, screening, due diligence, in the proximity of the final investment decision, in the contract stage, or subsequent rounds of investments (Aggarwal & Singh, 2013). Future research could dig deeper into these potential differences by leveraging finer-grained, qualitative data.

## 6. Conclusion

In this paper, we investigate the effect of media coverage on startups' VC funding. While extant research has investigated the importance of media coverage for startups' reputation and access to key resources, to the best of our knowledge, this is the first study to dive deeply into the characteristics of news to identify dimensions of media exposure uncovered by earlier works and examine their implications for startups' VC funding. We introduce a novel construct, media memorability, which combines three key dimensions that make a startup more attractive to stakeholders: prevalence, distinctiveness, and connectivity of a startup name in the semantic network of news. We hypothesize and test empirically the proposition that the combination of these three dimensions, as measured in news, has a significant effect on the memory of VCs. The findings based on survival analysis (and logit models as an additional test) confirm that the dimensions hypothesized in this paper, altogether, capture the attention of VCs and affect their decision to fund a startup. We also explore how the memorability effect on the probability of VC funding varies with media favorability. However, contrary to our expectations, our findings do not corroborate the hypothesis that media favorability reinforces the positive association between media memorability and VC funding, suggesting that investors value the two dimensions of media coverage independently. Previous work (Wei et al, 2017) has found that the interaction between favorability and media visibility negatively affects a firm's cumulative abnormal adjusted return (CAR), which highlights that the combined action of the two variables has complex, nuanced implications for investors' decision-making that future research could explore.

## Appendix

**Table A1**. List of media outlets of our sample

| Journal name | Number of news | Journal name | Number of news |
|---|---|---|---|
| The Journal (Newcastle, UK) | 131 | The Observer | 4 |
| Leicester Mercury | 90 | The Sentinel (Stoke) | 4 |
| Aberdeen Press and Journal | 73 | Yorkshire Evening Post | 4 |
| The Northern Echo | 57 | Evening Times (Glasgow) | 3 |
| The Scotsman | 56 | Mail on Sunday (London) | 3 |
| Belfast Telegraph | 51 | Metal Bulletin | 3 |
| Irish News | 50 | Scottish Business Insider | 3 |
| The Herald (Glasgow) | 47 | The Irish News | 3 |
| The Times (London) | 44 | Electronics Weekly | 2 |
| Belfast News Letter (Northern Ireland) | 43 | Herald Express (Torquay) | 2 |
| Evening News (Edinburgh) | 36 | Huddersfield Daily Examiner | 2 |
| Manchester Evening News | 35 | Hull Daily Mail | 2 |
| The Sunday Herald | 34 | Lancashire Evening Post | 2 |
| The Western Mail | 32 | Lancashire Telegraph | 2 |
| Daily Post (Liverpool) | 31 | Nottingham Evening Post | 2 |
| Evening Gazette | 30 | South Wales Echo | 2 |
| Yorkshire Post | 30 | Sunday Business | 2 |
| Business Telegraph | 29 | Sunday Telegraph (London) | 2 |
| UK Newsquest Regional Press - This is The NorthEast | 29 | The Sunday Telegraph (London) | 2 |
| Scotland on Sunday | 24 | World Oil | 2 |
| Independent on Sunday (London) | 22 | Banbury Guardian | 1 |
| Bath Chronicle | 20 | Bedfordshire on Sunday | 1 |
| Birmingham Post | 19 | Berwick Advertiser | 1 |
| Daily Mail (London) | 19 | Birmingham Evening Mail | 1 |
| The Express | 15 | Business 7 (UK) | 1 |
| The Times Higher Education Supplement | 15 | Carmarthen Journal | 1 |
| Aberdeen Evening Express | 14 | Corporate Financing Week | 1 |
| The Guardian (London) | 13 | Coventry Evening Telegraph | 1 |
| Evening Chronicle (Newcastle, UK) | 10 | Derby Evening Telegraph | 1 |
| Evening Herald (Plymouth) | 10 | Evening Standard (London) | 1 |
| The Evening Standard (London) | 9 | Express & Echo (Exeter) | 1 |
| Western Morning News (Plymouth) | 9 | Financial Post | 1 |
| UK Newsquest Regional Press - This is Lancashire | 8 | Financial Times | 1 |
| Western Daily Press | 8 | Gloucestershire Echo | 1 |
| Cambridge Evening News | 7 | Kidderminster Shuttle | 1 |
| Daily Record | 7 | Pontefract & Castleford Express | 1 |
| The Mirror | 7 | Sunday Business Post | 1 |
| UK Newsquest Regional Press - This is Bradford | 7 | Sunday Express | 1 |
| Sunday Times (London) | 6 | Sunday Mail | 1 |
| The Star (Sheffield) | 6 | Sunday Mirror | 1 |
| The Sunday Times (London) | 6 | The Engineer | 1 |
| UK Newsquest Regional Press - This is Wiltshire | 6 | The People | 1 |
| UK Newsquest Regional Press - This is Cheshire | 5 | The Sun (England) | 1 |
| UK Newsquest Regional Press - This is Oxfordshire | 5 | UK Newsquest Regional Press - This is Brighton and Hove | 1 |
| Bristol Evening Post | 4 | UK Newsquest Regional Press - This is Gwent | 1 |
| Daily Star | 4 | UK Newsquest Regional Press - This is Hampshire | 1 |
| South Wales Evening Post | 4 | | |

**Table A2.** Logit for the likelihood of receiving VC funding

| | (1) | (2) | (3) | (4) | (5) | (6) | (7) | (8) | (9) |
|---|---|---|---|---|---|---|---|---|---|
| London | 0.0652 | 0.0908 | 0.0892 | 0.0749 | 0.0905 | 0.0907 | 0.0900 | 0.0753 | 0.0907 |
| | (0.0930) | (0.0929) | (0.0929) | (0.0925) | (0.0928) | (0.0929) | (0.0928) | (0.0927) | (0.0928) |
| Team heterogeneity | 0.301** | 0.326** | 0.330** | 0.306** | 0.328** | 0.326** | 0.330** | 0.305** | 0.328** |
| | (0.130) | (0.134) | (0.134) | (0.132) | (0.134) | (0.134) | (0.134) | (0.132) | (0.134) |
| Academic prevalence | -0.0393 | -0.0208 | -0.0245 | -0.0443 | -0.0221 | -0.0209 | -0.0236 | -0.0405 | -0.0219 |
| | (0.142) | (0.142) | (0.145) | (0.145) | (0.143) | (0.142) | (0.145) | (0.145) | (0.143) |
| Patents | 0.543*** | 0.551*** | 0.551*** | 0.539*** | 0.550*** | 0.551*** | 0.551*** | 0.540*** | 0.550*** |
| | (0.0811) | (0.0799) | (0.0799) | (0.0802) | (0.0799) | (0.0797) | (0.0794) | (0.0806) | (0.0796) |
| Publications | 0.130 | 0.0796 | 0.0781 | 0.112 | 0.0788 | 0.0794 | 0.0790 | 0.116 | 0.0791 |
| | (0.121) | (0.115) | (0.116) | (0.123) | (0.115) | (0.115) | (0.116) | (0.123) | (0.115) |
| Academic spinoff | 0.127 | 0.101 | 0.101 | 0.114 | 0.100 | 0.101 | 0.101 | 0.111 | 0.101 |
| | (0.154) | (0.155) | (0.155) | (0.154) | (0.155) | (0.155) | (0.155) | (0.154) | (0.155) |
| Topic 1 – Public engagement | -0.205 | -0.290 | -0.270 | -0.229 | -0.284 | -0.290 | -0.276 | -0.234 | -0.285 |
| | (0.213) | (0.200) | (0.201) | (0.209) | (0.200) | (0.201) | (0.202) | (0.206) | (0.201) |
| Topic 2 – Product & Technology | 0.197 | 0.208 | 0.177 | 0.207 | 0.198 | 0.207 | 0.183 | 0.215 | 0.199 |
| | (0.237) | (0.203) | (0.198) | (0.219) | (0.200) | (0.204) | (0.200) | (0.218) | (0.202) |
| Topic 3 – Business growth | 0.147 | 0.164 | 0.179 | 0.144 | 0.169 | 0.165 | 0.176 | 0.137 | 0.168 |
| | (0.174) | (0.148) | (0.146) | (0.161) | (0.147) | (0.149) | (0.146) | (0.160) | (0.147) |
| Topic 4 – Management team | -0.0355 | -0.0121 | -0.0208 | -0.0365 | -0.0154 | -0.0121 | -0.0200 | -0.0355 | -0.0154 |
| | (0.153) | (0.0890) | (0.112) | (0.146) | (0.0961) | (0.0886) | (0.111) | (0.139) | (0.0965) |
| Topic 5 – Financial market | 0.0194 | 0.0225 | 0.0247 | 0.0256 | 0.0236 | 0.0226 | 0.0242 | 0.0263 | 0.0235 |
| | (0.0661) | (0.0565) | (0.0564) | (0.0611) | (0.0563) | (0.0563) | (0.0564) | (0.0609) | (0.0562) |
| Media favorability | 0.384*** | 0.348*** | 0.348*** | 0.374*** | 0.348*** | 0.347*** | 0.355*** | 0.388*** | 0.350*** |
| | (0.0849) | (0.0848) | (0.0852) | (0.0839) | (0.0850) | (0.0919) | (0.0888) | (0.0839) | (0.0909) |
| Prevalence | | 0.306*** | | | | 0.304*** | | | |
| | | (0.0951) | | | | (0.0740) | | | |
| Distinctiveness | | | 0.362*** | | | | 0.379*** | | |
| | | | (0.101) | | | | (0.0927) | | |
| Connectivity | | | | 0.263*** | | | | 0.321*** | |
| | | | | (0.0912) | | | | (0.0994) | |
| Media memorability | | | | | 0.335*** | | | | 0.340*** |
| | | | | | (0.106) | | | | (0.0852) |
| Prevalence*Favorability | | | | | | 0.004 | | | |
| | | | | | | (0.078) | | | |
| Distinctiveness*Favorability | | | | | | | -0.0240 | | |
| | | | | | | | (0.070) | | |
| Connectivity*Favorability | | | | | | | | -0.090* | |
| | | | | | | | | (0.051) | |
| Media memorability*Favorability | | | | | | | | | -0.008 |
| | | | | | | | | | (0.083) |
| Constant | -2.767*** | -2.802*** | -2.811*** | -2.791*** | -2.806*** | -2.802*** | -2.811*** | -2.796*** | -2.805*** |
| | (0.147) | (0.150) | (0.150) | (0.148) | (0.150) | (0.151) | (0.150) | (0.148) | (0.151) |
| Observations | 197 | 197 | 197 | 197 | 197 | 197 | 197 | 197 | 197 |
| Wald chi-sq | 108.46*** | 111.20*** | 116.10*** | 108.01*** | 111.90*** | 124.91*** | 128.04*** | 143.91*** | 129.01*** |
| Pseudo R² | 0.146 | 0.166 | 0.170 | 0.157 | 0.168 | 0.166 | 0.170 | 0.159 | 0.168 |

*Standard errors are in parentheses*
*** p<.01, ** p<.05, * p<.1